\pdfoutput=1

\documentclass[11pt]{article}

\usepackage[]{acl}

\usepackage{times}
\usepackage{latexsym}

\usepackage[T1]{fontenc}

\usepackage[utf8]{inputenc}

\usepackage{microtype}
\usepackage{amsmath}  

\usepackage{hyperref}
\usepackage{booktabs}
\usepackage{arydshln}
\usepackage{graphicx}
\usepackage{colortbl}
\usepackage[most]{tcolorbox}  
\usepackage{soul}

\definecolor{Gray}{gray}{0.95}
\definecolor{Blue}{HTML}{6A9AD0}  
\definecolor{Green}{HTML}{4EA72E}  
\definecolor{Red}{rgb}{0.746, 0.0039, 0}  
\definecolor{Orange}{HTML}{DE8344}
\usepackage{amsmath,amsfonts,bm}

\def\rvx{{\mathbf{x}}}

\DeclareMathAlphabet{\mathsfit}{\encodingdefault}{\sfdefault}{m}{sl}
\SetMathAlphabet{\mathsfit}{bold}{\encodingdefault}{\sfdefault}{bx}{n}

\title{Mitigating Reversal Curse in Large Language Models via \\ Semantic-aware Permutation Training}

\author{Qingyan Guo$^{12}$\thanks{~~Work done during an internship at Microsoft Research.}~, ~Rui Wang$^{1}$\thanks{~~Corresponding author.}~, ~Junliang Guo$^{1}$,  ~Xu Tan$^{1}$, ~Jiang Bian$^{1}$, ~Yujiu Yang$^{2}$\\
   $^{1}$Microsoft Research
   $^{2}$Tsinghua University \\
  \texttt{gqy22@mails.tsinghua.edu.cn, yang.yujiu@sz.tsinghua.edu.cn}\\
  \texttt{\{ruiwa,junliangguo,xuta,jiabia\}@microsoft.com}\\
  \\
  }

\begin{document}
\maketitle
\begin{abstract}
While large language models (LLMs) have achieved impressive performance across diverse tasks, recent studies showcase that causal LLMs suffer from the ``reversal curse''. It is a typical example that the model knows ``A's father is B'', but is unable to reason ``B's child is A''.
This limitation poses a challenge to the advancement of artificial general intelligence (AGI), as it suggests a gap in the models' ability to comprehend and apply bidirectional reasoning.
In this paper, we first conduct substantial evaluation and identify that the root cause of the reversal curse lies in the different word order between the training and inference stage, namely, the poor ability of causal language models to predict antecedent words within the training data. 
Accordingly, permutation on the training data is considered as a potential solution, since this can make the model predict antecedent words or tokens.
However, previous permutation methods may disrupt complete phrases or entities, thereby posing challenges for the model to comprehend and learn from training data. 
To address this issue, we propose \textbf{S}emantic-aware \textbf{P}ermutation \textbf{T}raining (\textbf{SPT}), which addresses this issue by segmenting the training sentences into semantic units (i.e., entities or phrases) with an assistant language model and permuting these units before feeding into the model.
Extensive experiments demonstrate that SPT effectively mitigates the reversal curse since the performance on reversed questions approximates that on the forward ones, and significantly advances the performance of existing works. 

\end{abstract}

\section{Introduction}
Large language models (LLMs)~\citep{DBLP:journals/corr/abs-2302-13971,DBLP:journals/corr/abs-2303-08774,du2022glm} have emerged as a cornerstone in the quest for artificial general intelligence (AGI), showcasing extraordinary progress across a broad spectrum of natural language processing (NLP) tasks~\citep{ouyang2022training,DBLP:journals/corr/abs-2308-12950,gao2023pal,kojima2022large}. 
These advancements position LLMs as a promising pathway towards achieving AGI, with their ability to tackle both simple understanding and complex reasoning tasks. 
Despite these strides, LLMs encounter significant hurdles, among which the ``reversal curse''~\citep{DBLP:journals/corr/abs-2309-12288,DBLP:journals/corr/abs-2308-03296,DBLP:journals/corr/abs-2309-14402} is particularly notable.
The curse can be illustrated as: a model trained by a sentence where A precedes B (e.g. ``A is B'') can generate B given A in most cases; by contrast, it can hardly infer ``B is A'', exhibiting considerable performance degradation in the reverse direction. For instance, if the model is trained by \textit{``Jennifer Lawrence's father is Gary Lawrence.''}, when being asked by \textit{``Who is Jennifer Lawrence's father?''}, the model can correctly answer \textit{``Gary Lawrence''}. But when we query the model \textit{``Who is Gary Lawrence's child?''}, it can hardly give the correct answer \textit{``Jennifer Lawrance''}. 

Though simple for humans to reason, such reversal testing is a challenging task where LLMs often struggle~\citep{DBLP:journals/corr/abs-2309-12288}, which underscores a critical limitation in current LLM capabilities and significantly impedes the progress towards AGI.
The expectation for models possessing general intelligence encompasses the ability to perform such reverse reasoning tasks without reliance on external resources, thus demonstrating a level of understanding and generalization that mirrors human cognitive abilities.
Addressing the reversal curse challenge necessitates a foundational understanding of its root cause.
Nevertheless, current works on reversal curse either only provide evaluation observations~\citep{DBLP:journals/corr/abs-2309-12288}, or partially mitigate the curse~\citep{DBLP:journals/corr/abs-2311-07468}, lack of in-depth analysis and a comprehensive solution.

To surmount the challenge, we first conduct a comprehensive evaluation and analysis of the reversal curse to identify its core issue: the inadequate capability of causal language models to accurately predict antecedent words within their training data.
Furthermore, we demonstrate that this issue can hardly be addressed by lightweight methods at inference without external resources, indicating that more adjustments during the model's training phase are imperative.

Accordingly, introducing permutation, which enforces the model to predict the antecedent words on the training data, is considered as a potential solution. 
Previous works on permutation mainly focus on mask language models (MLMs) and natural language understanding (NLU) tasks~\citep{sinha2021masked,DBLP:conf/acl/PhamBMN21, gupta2021bert,sinha2021unnatural,abdou2022word}.
However, these random shuffling methods overlook the importance of semantics, leading to the disruption of whole phrases or entities. Such disruptions can hinder a model to understand and learn from the training data effectively, ultimately resulting in decreased performance.

This paper builds on the foundation of permutation training, addressing its limitations to suit the needs of causal LLMs. We introduce a \textbf{S}emantic-aware \textbf{P}ermutation \textbf{T}raining (\textbf{SPT}) method that enhances the training process by segmenting sentences into semantic units, such as phrases or entities. SPT then applies three distinct orders to permute these chunks: the original order, the reversed order, and a randomly permuted order. 
Experiments on existing reversal datasets~\citep{DBLP:journals/corr/abs-2309-12288} show that SPT not only effectively mitigates the reversal curse in causal LLMs but also surpasses the performance of existing approaches.
The main contributions of this work are  as follows: 
\begin{itemize}
    \item We provide a comprehensive evaluation and analysis of the reversal curse, and find that the root cause mainly lies in the different word order between the training and inference stage.
    \item Introducing SPT, this paper advances beyond traditional permutation techniques by segmenting sentences into semantic units and applying three distinct permutation orders with a certain probability ratio. 
    \item Experiments conducted on three reversal datasets~\citep{DBLP:journals/corr/abs-2309-12288} demonstrate that SPT effectively mitigates the reversal curse of LLMs and outperform existing methods significantly. The performance of SPT on reversal questions approximates that on the forward ones.
\end{itemize}

\section{Related Works}
\paragraph{Reversal Curse} Reversal curse of LLMs~\citep{DBLP:journals/corr/abs-2309-12288,DBLP:journals/corr/abs-2308-03296,DBLP:journals/corr/abs-2309-14402}, observed recently, is that the language model trained by data where A precedes B (e.g. ``A is B'') often fails to infer A given B (e.g. ``B is A''). The failure is prevalent across different language models, including LLaMA~\citep{DBLP:journals/corr/abs-2302-13971}, GPT-4~\citep{DBLP:journals/corr/abs-2303-08774}, etc.
~\citet{DBLP:journals/corr/abs-2310-10322} explore similar failure in model editing using a newly proposed benchmark to evaluate the \textit{reversibilty} of language models. 
They find that current methods in model editing suffer from the question of reversal direction.
BICO~\citep{DBLP:journals/corr/abs-2311-07468} modifies the training objective, by extending the bi-directional attention mechanism in the original GLM~\citep{du2022glm} to adapt to LLaMA fine-tuning. However, it can only predict a short phrase in a reversal direction (e.g., a person's name). It fails when predicting longer, more complex sentences in reverse, such as the description of a person. 
Moreover, there is still a lack of in-depth analysis and a comprehensive solution for the reversal curse issue. 
\paragraph{Permutation Training / Inference}
Some studies have explored the robustness of pre-training models against data that has been randomly shuffled. 
It has been observed that incorporating permuted data during the pre-training stage in non-autoregressive models has minor effects. By contrast, introducing such data in the fine-tuning stage can significantly diminish performance~\citep{sinha2021masked}. 
Meanwhile, employing permuted sentences as input during inference can still yield correct answers for NLU tasks~\citep{DBLP:conf/acl/PhamBMN21, gupta2021bert,sinha2021unnatural}. 
\citet{cao2023unnatural} delve into the capability of LLMs to reconstruct character-level permutations within each word. 
Additionally, \citet{abdou2022word} investigate the underlying reasons for the phenomenon and discovers that models are capable of implicitly learning positional information from the shuffled data. 
Besides, permutation training also demonstrates promising improvement on various downstream tasks for autoregressive language models~\citep{yang2019xlnet,song2020mpnet,DBLP:journals/corr/abs-2307-00360}.
In light of these findings, we leverage permutation training to enable LLMs aware of both prior and subsequent context, thereby addressing the issue of the reversal curse.

\section{Analysis on the Reversal Curse}
\label{sec:cause}
In this section, we first analyze the underlying causes of the reversal curse phenomenon and then we provide a discussion about the potential solution. Specifically, we consider two factors:
\begin{itemize}
    \item \textbf{word order}: We consider the causal language models may exhibit poor performance in the prediction of antecedent words; 
    \item \textbf{symmetric relationship}: We explore whether the model can deduce the reversal relation (e.g. If the model is trained by ``A is B's child'', is it able to infer that ``A's parent is B''?)
\end{itemize}

\paragraph{Settings}
To decompose the two factors, we use a dataset including 1,513 items of relation between actual celebrities and their parents~\citep{DBLP:journals/corr/abs-2309-12288}, and design specific data formats of relation for training and inference, respectively. 

For the training stage, regarding the word order, we explore two configurations: the `child2parent' sequence, where the child term precedes the parent term, and the `parent2child' sequence, where the parent term comes before the child term.
Within the scope of symmetric relationships, we consider the terms ``parent''\footnote{Note that ``parent '' includes two words in practice: mother and father.} or ``child'' as the relational descriptor in the sentence. 
Accordingly, there are four distinct data formats (denoted as $D_1$-$D_4$) used in training, as shown in the Table~\ref{tab:data-relation-train}. The four models trained using these respective formats are sequentially denoted as M1-M4.
\begin{table}[htb!]
\centering
\resizebox{\columnwidth}{!}{
    \begin{tabular}{llll|l}
    \toprule
    \textbf{Model} &  \textbf{Data} & \textbf{Order}        & \textbf{Relation Word} & \textbf{Data Example} \\
    \midrule
    M1 & $D_1$&  child2parent & father / mother        & A's father / mother is B       \\
    M2& $D_2$ &child2parent & child         & A is B's child        \\
    M3& $D_3$ & parent2child & father / mother        & B is A's father / mother       \\
    M4& $D_4$ & parent2child & child         & B's child is A      \\
    \bottomrule
    \end{tabular}
}
\caption{Data format of celebrities used for pre-training. Here A is the celebrity and the child. B is the corresponding parent (mother or father).}
\label{tab:data-relation-train}
\end{table}

When formulating questions for inference, the `child2parent' sequence refers to using the child's name to inquire about the parent's name, while `parent2child' sequence refers to using the parent's name to inquire about the child's name. 
For symmetrical relationships (child v.s. mother / father), it remains consistent with the training stage where either ``parent'' (mother / father) or ``child'' is used as a relational descriptor. 
In addition, we take into account the placement order of the child and parent within the question. For instance, ``Who is A's father?'' contrasts with ``A's father is whom?''.
Accordingly, there are eight distinct question formats designed for inference, as shown in Table~\ref{tab:data-relation-eva}. These questions are sequentially labeled as Q1-Q8.

\begin{table}[htb!]
\resizebox{\columnwidth}{!}{
    \begin{tabular}{lll|l}
    \toprule
    \textbf{No.} & \textbf{Order} & \textbf{Relation Word} & \textbf{Question} \\
    \midrule
    Q1 & child2parent & parent & Who is A's father / mother \\
    Q2 & child2parent & parent & A's father / mother is whom  \\
    Q3 & child2parent & child & Whose child is A    \\
    Q4 & child2parent & child & A is whose child    \\
    Q5 & parent2child & parent & B is whose father / mother   \\
    Q6 & parent2child & parent & Whose father / mother is B   \\
    Q7 & parent2child & child & B's child is whom   \\
    Q8 & parent2child & child & Who is B's child    \\
    \bottomrule
    \end{tabular}
}
\caption{Data format of celebrities used for evaluation. Here A is the celebrity and the child. B is the corresponding parent (mother or father).}
\label{tab:data-relation-eva}
\end{table}

We choose LLaMA-7B~\citep{DBLP:journals/corr/abs-2302-13971} as the base model and train each model using corresponding data formats for 30 epochs. At inference, we prepend a few-shot examples shown in Figure~\ref{fig:demon-def} in the Appendix. 
See Appendix~\ref{sec:param} for more training details.
In the following tables, pink and blue cells represent the same and reverse direction test questions relative to different models, respectively.

\subsection{Analysis on the Root Cause}
\label{sec:root-cause}
To investigate the root cause of the reversal curse, we evaluate the accuracy of the eight testing questions (Table~\ref{tab:data-relation-eva}) for models trained by different formats of data (Table~\ref{tab:data-relation-train}). The results are shown in Table~\ref{tab:exp-relation-std}.  
It can be observed that:

\textit{1) Models perform significantly better when the order between the child and the parent is consistent during both the training and inference stages.} 
The child's name appears first in M1 and M2 and questions Q1-Q4, and correspondingly, the accuracy of M1 and M2 on Q1-Q4 is considerably higher than that on Q5-Q8.
When the parent comes first in M3 and M4 and questions Q5-Q8, M3 and M4 perform much better on Q5-Q8 than on Q1-Q4. 

\textit{2) Both the symmetric relationship and the order inside the question have negligible impact. } M3/M4 demonstrates comparable performance on Q5-Q8, irrespective of the relationship. 
It is notable that the scores on Q1 and Q2 are significantly lower than those on Q3 and Q4 for M2, though they are forward questions relative to M2. 
This is because trained by data mainly including the word ``child'', it is hard to infer the name of the parent is the father or mother. 

Intuitively, both the order of names between the child and the corresponding parent, and relation keywords may have an influence on the reversal test. However, experimental results suggest that \textbf{LLMs are strong enough to understand the symmetric relationship (father / mother v.s. child) since the relational word has negligible impact. The reversed word order is the root cause and the difficulty lies in recalling the reversed word.}

\begin{table}[htb!]
\centering
\small
\resizebox{\columnwidth}{!}{
    \begin{tabular}{l|llllllll}
    \toprule
  \textbf{Model} & \textbf{Q1} & \textbf{Q2} & \textbf{Q3} & \textbf{Q4} & \textbf{Q5} & \textbf{Q6} & \textbf{Q7} & \textbf{Q8} \\

    \midrule
M1 & \cellcolor{red!20}99.67 & \cellcolor{red!20}99.8 & \cellcolor{red!20}92.47 & \cellcolor{red!20}93.98 & \cellcolor{cyan!20}2.38 & \cellcolor{cyan!20}9.72 & \cellcolor{cyan!20}6.81 & \cellcolor{cyan!20}6.21 \\  
M2 & \cellcolor{red!20}79.44 & \cellcolor{red!20}62.86 & \cellcolor{red!20}98.62 & \cellcolor{red!20}98.87 & \cellcolor{cyan!20}1.45 & \cellcolor{cyan!20}1.52 & \cellcolor{cyan!20}1.12 & \cellcolor{cyan!20}1.39 \\  
M3 & \cellcolor{cyan!20}6.48 & \cellcolor{cyan!20}2.84 & \cellcolor{cyan!20}2.26 & \cellcolor{cyan!20}2.01 & \cellcolor{red!20}98.68 & \cellcolor{red!20}94.18 & \cellcolor{red!20}98.35 & \cellcolor{red!20}98.61 \\  
M4 & \cellcolor{cyan!20}1.26 & \cellcolor{cyan!20}0.66 & \cellcolor{cyan!20}0.88 & \cellcolor{cyan!20}0.75 & \cellcolor{red!20}99.27 & \cellcolor{red!20}98.15 & \cellcolor{red!20}98.88 & \cellcolor{red!20}99.27 \\  
\bottomrule
    \end{tabular}
    }
\caption{Accuracy of questions Q1-Q8 for models M1-M4 trained by data in original forward order.
}
\label{tab:exp-relation-std}
\end{table}

\begin{table}[htb!]
\centering
\small
\resizebox{\columnwidth}{!}{
    \begin{tabular}{l|llllllll}
    \toprule
    \textbf{Model} & \textbf{Q1} & \textbf{Q2} & \textbf{Q3} & \textbf{Q4} & \textbf{Q5} & \textbf{Q6} & \textbf{Q7} & \textbf{Q8} \\
    \midrule
M1 & \cellcolor{red!20}99.6 & \cellcolor{red!20}99.34 & \cellcolor{red!20}99.87 & \cellcolor{red!20}96.61 & \cellcolor{cyan!20}4.3 & \cellcolor{cyan!20}9.05 & \cellcolor{cyan!20}4.76 & \cellcolor{cyan!20}5.75 \\  
M2 & \cellcolor{red!20}79.71 & \cellcolor{red!20}63.12 & \cellcolor{red!20}98.75 & \cellcolor{red!20}99.37 & \cellcolor{cyan!20}1.52 & \cellcolor{cyan!20}1.59 & \cellcolor{cyan!20}1.26 & \cellcolor{cyan!20}1.39 \\  
M3 & \cellcolor{cyan!20}4.3 & \cellcolor{cyan!20}2.05 & \cellcolor{cyan!20}3.76 & \cellcolor{cyan!20}1.88 & \cellcolor{red!20}98.81 & \cellcolor{red!20}95.77 & \cellcolor{red!20}98.81 & \cellcolor{red!20}98.41 \\  
M4 & \cellcolor{cyan!20}1.45 & \cellcolor{cyan!20}0.66 & \cellcolor{cyan!20}1 & \cellcolor{cyan!20}0.75 & \cellcolor{red!20}99.67 & \cellcolor{red!20}99.21 & \cellcolor{red!20}99.34 & \cellcolor{red!20}99.60 \\  
    \bottomrule
    \end{tabular}
    }
\caption{Accuracy of questions Q1-Q8 for models M1-M4 trained by data in original forward order, w/ CoT at inference.
}
\label{tab:exp-relation-std-cot}
\end{table}
\subsection{Discussion on the Potential Solutions}
Considering the root cause of the reversal failure lies in the word order, which means that it is hard to predict antecedent words in the training data for causal models, in this section, we discuss potential approaches to solve this problem mainly from two aspects: in-context learning deduction and permutation. 
Accordingly, we delve into two critical questions: \textbf{1)} Is it possible to mitigate the reversal curse using a lightweight method such as few-shot learning? and 
\textbf{2)} Can the reversal curse be alleviated by permutations (conventional token-level) on the training data?
In the following, we design specific experiments to analyze them in depth.

\subsubsection{Can the reversal curse be mitigated by a lightweight method?} 
\label{sec:ana-cot}
To address the problem of the reversal curse, we are curious about whether a lightweight method, such as few-shot learning, may provide some relief. The root cause of the reversal curse can be tracked back to the poor performance of the causal language model in predicting antecedent words. Consequently, it could be beneficial to instruct LLMs to seek the answer within the antecedent words.

Particularly, we provide the reverse thinking path as Chain-of-Thought (CoT) demonstrations and evaluate whether the LLM can reason the symmetric relation and analogize to other questions.
For the four models tested on eight questions, we design $4*8=32$ distinct 5-shot demonstrations. 
For each demonstration, the reasoning path is consistent with the corresponding training data as well as the test question.
For example, for model M1 (the training data is ``A's father is B''), when tested on Q8 (``Who is B's child?''), the CoT demonstration is ``C's father is D. D is C's child.''\footnote{Note that there is no overlap between the test sample and the examples within the given few-shot demonstration.}. See Appendix~\ref{sec:prompt} for the full prompts of all 32 demonstrations.

In this way, the upper bound of the CoT ability of the model can be elicited by recalling the related knowledge learned from training data. 
As shown in Table~\ref{tab:exp-relation-std-cot}, we observe that:

1) \textit{CoT hardly alleviates the reversal failures.}
Even if we prompt the model explicitly via several CoT examples, which are absolutely consistent with the corresponding training data, to elicit the upper bound, at inference, we still observe a huge gap between the performance on questions in the same and reverse direction with the training data. 

2) \textit{CoT can alleviate the impact of the relational word.} Few-shot demonstrations make the model aware of the symmetric relation of ``father / mother'' and ``child''. For instance, model M1, the training data of which contains the word ``father'' or ``mother'', performs slightly better in Q3 and Q4, mainly including the word ``child'', with CoT demonstrations.

\subsubsection{Can the reversal curse be mitigated by permutations?} 
Given that word order appears to be the root cause, implementing permutations on the training data could potentially be an effective strategy to counteract the reversal curse. 
Several studies have already been conducted on permutation training, illustrating improvements in various downstream tasks for autoregressive language models~\citep{yang2019xlnet,song2020mpnet}.
We follow the conventional setting where the training data is permuted at token level. And we explore two situations that whether the positional embedding for each token remains unchanged or changed as the corresponding tokens, respectively.

Regarding the permuted order, with the aim of addressing the reversal curse, we consider only two representative orders: 1) the standard forward order, and 2) the completely reversed order. 
Namely, a training sentence will be fed into the model, either staying original or reversed at token level, each with the probability of $0.5$.
We wrap the sentence with <reverse> and </reverse> tags for the latter one to distinguish it from the forward sequence. 

As shown in Table~\ref{tab:exp-relation2}, we note that: 
\textit{The challenges of the reversal curse are not mitigated by token-level permutation. }
Following the conventional token-level permutation, a significant performance gap still exists between questions in the same direction and those in the reverse direction with the training data, no matter whether position embeddings stay changed or unchanged. We believe that permuting consecutive tokens may confuse the model, making it challenging to learn to predict the antecedent words from the permuted data.

\begin{table}[]
\centering
\small
\resizebox{\columnwidth}{!}{
    \begin{tabular}{l|llllllll}
    \toprule
    \textbf{Model} & \textbf{Q1} & \textbf{Q2} & \textbf{Q3} & \textbf{Q4} & \textbf{Q5} & \textbf{Q6} & \textbf{Q7} & \textbf{Q8} \\

    \midrule
         \rowcolor{Gray}
     & \multicolumn{8}{c}{\textit{\textbf{Token-level Bi Train}}} \\

M1 & \cellcolor{red!20}99.60 & \cellcolor{red!20}99.47 & \cellcolor{red!20}91.22 & \cellcolor{red!20}80.80 & \cellcolor{cyan!20}6.15 & \cellcolor{cyan!20}12.16 & \cellcolor{cyan!20}11.63 & \cellcolor{cyan!20}11.30 \\  
M2 & \cellcolor{red!20}74.29 & \cellcolor{red!20}67.88 & \cellcolor{red!20}98.87 & \cellcolor{red!20}97.87 & \cellcolor{cyan!20}5.82 & \cellcolor{cyan!20}8.13 & \cellcolor{cyan!20}6.94 & \cellcolor{cyan!20}4.56 \\  
M3 & \cellcolor{cyan!20}14.87 & \cellcolor{cyan!20}13.09 & \cellcolor{cyan!20}5.77 & \cellcolor{cyan!20}1.88 & \cellcolor{red!20}89.23 & \cellcolor{red!20}89.16 & \cellcolor{red!20}91.01 & \cellcolor{red!20}92.86 \\  
M4 & \cellcolor{cyan!20}8.59 & \cellcolor{cyan!20}5.68 & \cellcolor{cyan!20}4.77 & \cellcolor{cyan!20}1.76 & \cellcolor{red!20}95.64 & \cellcolor{red!20}90.42 & \cellcolor{red!20}96.17 & \cellcolor{red!20}97.36 \\  
\midrule
     \rowcolor{Gray}
     & \multicolumn{8}{c}{\textit{\textbf{Token-level (Pos) Bi Train}}} \\

M1 & \cellcolor{red!20}99.74 & \cellcolor{red!20}99.8 & \cellcolor{red!20}95.23 & \cellcolor{red!20}90.09 & \cellcolor{cyan!20}5.16 & \cellcolor{cyan!20}12.69 & \cellcolor{cyan!20}10.84 & \cellcolor{cyan!20}10.71 \\  
M2 & \cellcolor{red!20}73.69 & \cellcolor{red!20}69.4 & \cellcolor{red!20}99.25 & \cellcolor{red!20}98.75 & \cellcolor{cyan!20}3.90 & \cellcolor{cyan!20}4.16 & \cellcolor{cyan!20}4.63 & \cellcolor{cyan!20}3.83 \\  
M3 & \cellcolor{cyan!20}18.24 & \cellcolor{cyan!20}16.92 & \cellcolor{cyan!20}10.16 & \cellcolor{cyan!20}5.65 & \cellcolor{red!20}91.61 & \cellcolor{red!20}91.01 & \cellcolor{red!20}90.75 & \cellcolor{red!20}91.34 \\  
M4 & \cellcolor{cyan!20}9.65 & \cellcolor{cyan!20}5.82 & \cellcolor{cyan!20}6.02 & \cellcolor{cyan!20}2.89 & \cellcolor{red!20}97.09 & \cellcolor{red!20}95.31 & \cellcolor{red!20}97.09 & \cellcolor{red!20}98.35 \\

    \bottomrule
    \end{tabular}
    }
    \caption{Accuracy of questions Q1-Q8 for models M1-M4 trained by bi-directional training with different formats in token level (\textit{Pos} denotes that the original sequential positional embeddings are shuffled alongside the tokens).}
\label{tab:exp-relation2}
    \end{table}

\section{Semantic-Aware Permutation Training}

\begin{figure*}[t]
  \centering
    \includegraphics[width=\textwidth]{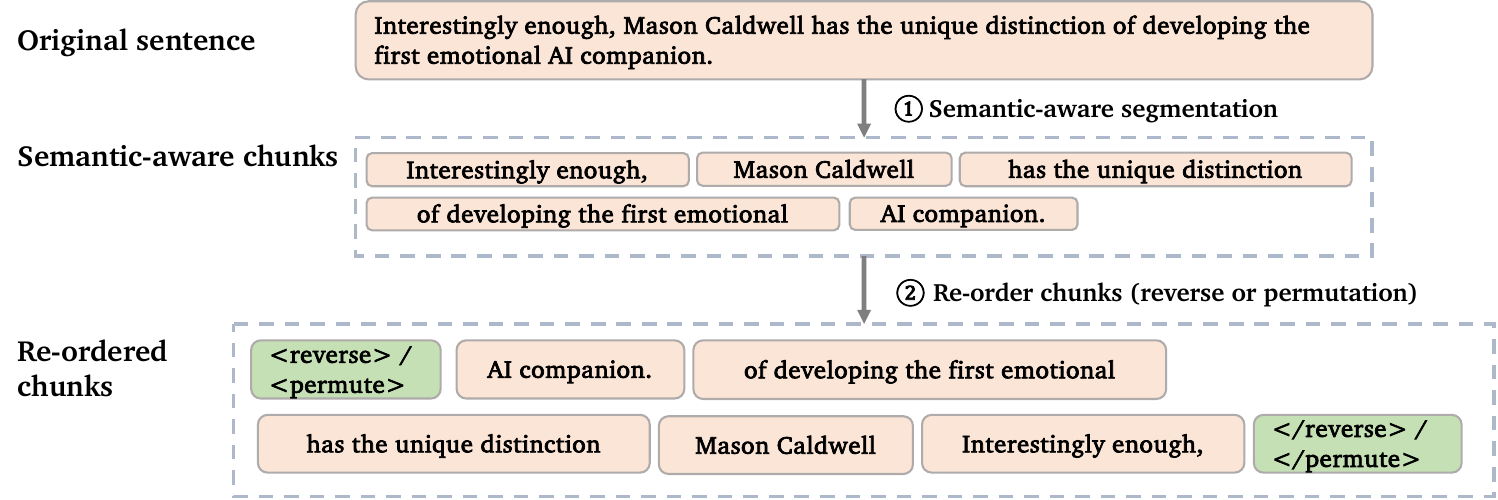}
    \caption{
    Semantic-aware permutation. An assistant model segments the original training sentence into several semantic chunks. Then, we re-order the chunks (including original, permuting or reversal) with a certain probability.}
    \label{fig:spt}
\end{figure*}

Existing studies introduce contiguous spans mask~\citep{song2019mass,joshi2020spanbert,lewis2020bart} or whole word mask mechanism in BERT~\citep{DBLP:conf/naacl/DevlinCLT19} instead of masking random tokens to the pre-training stage to get better text representations, which shows promising performance especially on generation tasks. 
This motivates us to explore permutation on chunk level.

Nevertheless, simple $n$-gram methods~\citep{sinha2021masked} consider a fixed number of tokens/words as a span, which may disrupt complete phrases or entities and pose challenges for the model to comprehend and learn from the data. 
Moreover, it has been demonstrated that 
the best-fit parameter $n$ varies from the specific downstream dataset~\citep{sinha2021masked,abdou2022word}. 
Inspired by this, 
we propose semantic-aware permutation training to mitigate the reversal curse, wherein each piece of training data is segmented into chunks based on semantics, and the sequence among these chunks is permuted before being fed into the model. Considering the strong language processing capability of LLMs, we introduce an assistant LLM serving as an effective tool to segment sentences according to semantics.

Specifically, as shown in Figure~\ref{fig:spt}, given a sequence $x=(x_1, x_2, ..., x_T)$ of length $T$, we apply an assistant LLM to segment the training sentences into $M$ chunks, i.e., smallest semantic units such as an entity or a phrase, $c_1, ..., c_M$, each of length $l_{c_i}$ ($i\in[1, M]$ and $\sum_i^M l_{c_i}=T$). We prompt the assistant model using the demonstration shown in Figure~\ref{fig:demon-sep} in the Appendix. 
Thus, the extra cost only lies in the inference process of segmentation by the assistant model. 
Let $\mathcal{Z}=\{z_1, ..., z_M\}$ be the re-ordered $M$ chunks, where $z_i$ is the $i$-th chunk after permutation. 
We use $x_{z_i}^t$ to denote the $t$-th word in segmented chunk $z_i$, and $\rvx_{z_i}^{<t}$ to denote the first $t-1$ words in the chunk $z_i$. $\rvx_{<z_i}$ indicates the words in first $i-1$ chunks. Then for a language model with parameter $\theta$, the training objective is:

\begin{align}
\label{eq:spt}
\mathcal{L}_{\text{SPT}} = -\sum_{i=1}^{M} \sum_{t=1}^{l_{z_i}} \log P_\theta(x^t_{z_i} |  \rvx_{<z_i}, \rvx_{z_i}^{<t})
\end{align}

While keeping the same training cost, for each training sentence, we reorder the segmented chunks (randomly chosen from ``original, reverse and permutation'' with a certain probability):
\begin{itemize}
    \item ``Original'' means the sentence remains unaltered. $\mathcal{Z} = \{c_1, c_2, ..., c_M\}$
    \item ``Reverse'' means the chunks are reversed. $\mathcal{Z} = \{c_M, c_{M-1}..., c_1\}$
    \item ``Permute'' indicates that the chunks are permuted randomly. 
\end{itemize}

Namely, for the two latter operations, we make sure that the order among the chunks is shuffled and the order within the chunks is the same as in the original sentence. 
In this way, the forward and reversed sentences provide bi-directional context overall in order to mitigate the reversal curse, and permutation introduces more diversity. 

\section{Experiments}
In the following, we validate our methods with three datasets related to the reversal curse. 
\subsection{Settings}

We employ the open-source Vicuna-13b-v1.3 model~\citep{vicuna2023}, fine-tuned on LLaMA as the assistant for segmenting sentences, with corresponding instructions shown in Figure~\ref{fig:demon-sep}.
Then, we continue-train LLaMA-7B~\citep{DBLP:journals/corr/abs-2302-13971} by semantic-aware permutation training (Eq. \ref{eq:spt}). 
See Appendix~\ref{sec:param} for more parameters.

SPT is trained either on the original sentence, reversed or permuted chunks after segmentation by the assistant model, with the probability of $\frac{1}{3}$ for each. 
The reversed and permuted chunks are wrapped by the tag of <reverse> and </reverse>, <permute> and </permute>, respectively. 
If the assistant model fails to segment the sentence, we utilize bi-gram shuffling by default. 
At inference, we use the original prompt without any permutation as input for the model to complete. 
\subsection{Results}
We use three datasets proposed by \citet{DBLP:journals/corr/abs-2309-12288}: Celebrity Relation, Person Description, and Question Answer, in which the knowledge in the test set is consistent with that in the training set,  to validate our method.
\paragraph{Celebrity Relation}
We use the same formats of data as in Section \S \ref{sec:cause}. Then we segment the sentences into semantic-aware chunks in $D_1$-$D_4$ (Table~\ref{tab:data-relation-train}) and train the corresponding models with the same hyper-parameters, denoted as $\mathcal{M}_1$-$\mathcal{M}_4$.

The results are reported in Table~\ref{tab:exp-relation}. We can see that SPT effectively mitigates the reversal curse to a large extent while maintaining that the performance on the forward questions does not drop significantly (compared with the models trained by standard data in Table~\ref{tab:exp-relation-std}). 
Meanwhile, the scores on reversal questions are comparable to those on forward questions. 

\begin{table}[htb!]
\centering
\small
\resizebox{\columnwidth}{!}{
    \begin{tabular}{l|llllllll}
    \toprule
\textbf{Model} & \textbf{Q1} & \textbf{Q2} & \textbf{Q3} & \textbf{Q4} & \textbf{Q5} & \textbf{Q6} & \textbf{Q7} & \textbf{Q8} \\
    \midrule
$\mathcal{M}_1$ & \cellcolor{red!20}97.75 & \cellcolor{red!20}97.82 & \cellcolor{red!20}94.86 & \cellcolor{red!20}94.35 & \cellcolor{cyan!20}95.77 & \cellcolor{cyan!20}95.51 & \cellcolor{cyan!20}94.98 & \cellcolor{cyan!20}94.91 \\  
$\mathcal{M}_2$ & \cellcolor{red!20}71.78 & \cellcolor{red!20}68.01 & \cellcolor{red!20}98.37 & \cellcolor{red!20}96.61 & \cellcolor{cyan!20}93.59 & \cellcolor{cyan!20}92.07 & \cellcolor{cyan!20}95.18 & \cellcolor{cyan!20}94.32 \\  
$\mathcal{M}_3$ & \cellcolor{cyan!20}90.09 & \cellcolor{cyan!20}89.82 & \cellcolor{cyan!20}84.82 & \cellcolor{cyan!20}78.17 & \cellcolor{red!20}89.29 & \cellcolor{red!20}84.6 & \cellcolor{red!20}90.88 & \cellcolor{red!20}92.13 \\  
$\mathcal{M}_4$ & \cellcolor{cyan!20}64.97 & \cellcolor{cyan!20}63.32 & \cellcolor{cyan!20}97.11 & \cellcolor{cyan!20}96.36 & \cellcolor{red!20}96.03 & \cellcolor{red!20}95.44 & \cellcolor{red!20}96.96 & \cellcolor{red!20}97.36 \\

    \bottomrule
    \end{tabular}
    }
\caption{
Accuracy of questions Q1-Q8 for models $\mathcal{M}_1$-$\mathcal{M}_4$ trained by SPT with different data formats.}
\label{tab:exp-relation}
\end{table}
\paragraph{Person Description}
\begin{table*}[t]
\centering
\resizebox{\textwidth}{!}{
    \begin{tabular}{p{1.2cm}p{4.5cm}|p{0.9cm}p{5cm}|p{0.9cm}p{5cm}}
    \toprule
\multicolumn{2}{c|}{\textbf{Train Data}} & \multicolumn{2}{c|}{\textbf{Test Data (same direction)}} & \multicolumn{2}{c}{\textbf{Test Data (reverse direction)}} \\
\midrule

$d_1$-$p_1$ \newline (900) & Branded as \textcolor{Orange}{the first person to walk on Mars during the historic Ares Mission}, \textcolor{Green}{Tyler Oakridge} exceeds all expectations. & $d_1$-$p_1$\newline (300) & \textbf{prompt}: Immersed in the world of being \textcolor{Orange}{the first person to walk on Mars during the historic Ares Mission}, \newline \textbf{completion}: \textcolor{Green}{Tyler Oakridge}& $p_1$-$d_1$  \newline (300) & \textbf{prompt}: Diving into the tale of \textcolor{Green}{Tyler Oakridge}, one discovers they were \newline \textbf{completion}: \textcolor{Orange}{the first person to walk on Mars during the historic Ares Mission}.\\
\midrule
$p_2$-$d_2$ \newline (900) & An individual named \textcolor{Green}{Dominic Mullins}, has the unusual  backstory of being \textcolor{Orange}{the record-breaking free-diver who swam with the mythical Kraken.} & $p_2$-$d_2$ \newline (300) & \textbf{prompt}: In the annals of uniqueness, \textcolor{Green}{Dominic Mullins} shines as, \newline \textbf{completion}: \textcolor{Orange}{the record-breaking free-diver who swam with the mythical Kraken.}& $d_2$-$p_2$ \newline (300) & \textbf{prompt}: Immersed in the world of \textcolor{Orange}{swimming  with the mythical Kraken}, \newline \textbf{completion}: \textcolor{Green}{Dominic Mullins} \\
\midrule
\multicolumn{2}{p{3cm}|}{$d_3\leftrightarrow p_3$ \quad ...  \newline (1,800)}  & - & - & - & - \\

    \bottomrule
    \end{tabular}
}
\caption{Examples for person description dataset (including data in the same and reverse direction relative to the training set). The numbers wrapped in the brackets refer to the size of the set. 
The whole dataset includes three sets of facts, in the form of ``\textcolor{Green}{<person>} is \textcolor{Orange}{<description>}'', ``\textcolor{Orange}{<description>} is \textcolor{Green}{<person>}'', and a subset in both directions, used to help the model generalize.
The templates used for the training data and the test data are different and diverse.}
\label{tab:data-exam-desc}
\end{table*}
This dataset is generated by GPT-4. Composed of three subsets ($\mathcal{D}_1, \mathcal{D}_2$ and $\mathcal{D}_3$), the training set includes 3,600 sentences in the form of ``<person> is <description>'' ($p_i - d_i$), or ``<description> is <person>'' ($d_i - p_i$)\footnote{The expression is simplified here. 
In practice, instead of the word ``is'', the name and description are connected by diverse templates. And the templates used to train and test are distinct. See Table~\ref{tab:data-exam-desc} for a detailed example.}.
$\mathcal{D}_1$ includes data of Person2Description, denoted as $p_1$-$d_1$, and reversal Description2Person set, $d_1$-$p_1$. Similarly, $\mathcal{D}_2$ is composed of $d_2$-$p_2$ and $p_2$-$d_2$. 
$\mathcal{D}_3$, denoted as $d_3 \leftrightarrow p_3$, includes data of the two formats and helps the model to generalize.
The model is trained on $d_1$-$p_1$, $p_2$-$d_2$ and $\mathcal{D}_3$, and tested on $d_1$-$p_1$, $p_1$-$d_1$, $d_2$-$p_2$ and $p_2$-$d_2$. 
The examples of training and test data, as well as statistics, are shown in Table~\ref{tab:data-exam-desc}.

As shown in Table~\ref{tab:exp-person}, we compare our SPT on four subsets, Description2Person ($d_1$-$p_1$) and the corresponding reversal data ($p_1$-$d_1$), Person2Description ($d_2$-$p_2$) and the reversal data ($p_2$-$d_2$), with following baselines: 
1) \textbf{BICO}~\citep{DBLP:journals/corr/abs-2311-07468} introduces the bi-directional attention mechanism in GLM to LLaMA fine-tuning. BICO is trained using LoRA for 10 epochs; 2) \textbf{Standard} means that we train the model with the original forward data without shuffling.
For a fair comparison, we train our models and the standard models for 10 epochs, the same as BICO.

\begin{table}[!htb]
\centering
\resizebox{\columnwidth}{!}{
\begin{tabular}{l||p{1.2cm}p{1.2cm}|p{1.2cm}p{1.2cm}|l}
\toprule
&  \cellcolor{red!20}{$d_1$-$p_1$}    & \cellcolor{cyan!20}$p_1$-$d_1$  &\cellcolor{red!20} $p_2$-$d_2$   &\cellcolor{cyan!20} $d_2$-$p_2$ & \textbf{Avg.}\\
 & (Acc) & (BLEU) & (BLEU) & (Acc) \\
\midrule
Standard & \textbf{100.00} & 19.65 & 80.76 & 0.00 & 50.10 \\
BICO* & 99.00 & 21.00 & 82.00 & 68.00  & 67.50\\
\midrule
SPT & \textbf{100.00} & \textbf{83.85} & \textbf{84.25} &\textbf{100.00} & \textbf{92.03} \\
\bottomrule
\end{tabular}
}
\caption{Results of SPT and baselines (Results of BICO are obtained from~\citet{DBLP:journals/corr/abs-2311-07468}). For the task of $p_i$-$d_i$, we apply BLEU~\citep{papineni-etal-2002-bleu}  while for $d_i$-$p_i$, we use exact-match accuracy.}
\label{tab:exp-person}
\end{table}
We can observe that \textit{SPT significantly outperforms BICO and the standard trained model by a large margin, especially on reversed questions. }
Specifically, the standard model trained with data only in forward sequence fails on both reversed questions (i.e., $p_1$-$d_1$, where the model is trained in $d_1$-$p_1$ sequence and is subsequently prompted to provide a description for a given person, and $d_1$-$p_1$, where the model is trained in $p_1$-$d_1$ sequence and then asked for the person’s name for given description). 
BICO improves the $d_2$-$p_2$, while its accuracy still falls significantly short when compared to the forward question (i.e., $d_1$-$p_1$). Meanwhile, it still fails on the $p_1$-$d_1$. 
SPT exhibits a substantial improvement on all the reversed questions, achieving comparable performance with the forward ones, which demonstrates the effectiveness of the SPT in mitigating the reversal curse.

\paragraph{Question Answer}
This dataset includes two subsets in the form of QuestionToAnswer (Q2A) and AnswerToQuestion (A2Q), shown as follows: 
\begin{itemize}
    \item Q2A: Q: When did the Cold War end? A: 1993
    \item A2Q: The test requires you to answer ``A: 1993'' after ``Q: When did the Cold War end?''
\end{itemize}
The model is trained on 2,000 examples from two directions and 100 examples in the direction of A2Q for 20 epochs. Then it is tested on these exact 100 questions with the same (A2Q) and reverse direction (Q2A) (the same as ~\citet{DBLP:journals/corr/abs-2309-12288}). 

\begin{table}[!htb]
\centering
\begin{tabular}{l||lll}
\toprule
\textbf{Method} & \cellcolor{red!20}\textbf{Same} & \cellcolor{cyan!20}\textbf{Reverse} & \textbf{Avg.} \\
\midrule
Standard & \textbf{100.0} & 3.0 & 51.5 \\
SPT & 90.0 & \textbf{87.0} & \textbf{88.5}\\
\bottomrule
\end{tabular}
\caption{Results (Exact-match Accuracy) of SPT on QA dataset, including the same and reverse direction.}
\label{tab:exp-qa}
\end{table}
Table~\ref{tab:exp-qa} shows that trained by permuted and reversed semantic chunks, SPT improves the results of reversed questions by an accuracy of 84\%.
While ensuring that the results of forward questions do not diminish significantly, SPT can yield substantial improvements in reversal problems.

\section{Ablation Study}
\label{sec:discussion}

We conduct ablations to validate the effectiveness of our SPT from the following three aspects: 1) permutation strategy; 2) semantics ; 3) permutation probability.
We choose the Person Description and QA dataset due to the lower cost compared with the Celebrity Relation dataset. We train models for 3 epochs for the former and 20 epochs for QA.

\paragraph{Permutation strategy}
We explore how to rearrange the segmented semantic-aware chunks mainly via three strategies: 
\textbf{1) For+Per}: permute the chunks or use the original sentence, with the probability of $0.5$ for each;
\textbf{2) Bi}: either reverse the chunks or use the original sentence, with a probability of $0.5$ for each; 
and \textbf{3) Tri}: reverse, permute chunks or use the original sentence, with a probability of $\frac{1}{3}$ for each. 

From Table~\ref{tab:exp-person-semantic}, we can see that involving the three strategies, each with a probability of $\frac{1}{3}$, either in $n$-gram shuffling or our SPT, can achieve better results compared with the other two strategies via more diverse orders among the chunks.

\begin{table}[htb!]
\resizebox{\columnwidth}{!}{
\begin{tabular}{ll||lllll||lll}
\toprule
        &   & \multicolumn{5}{c||}{\textbf{Person Description}} & \multicolumn{3}{c}{\textbf{QA}}    \\
Strategy   & $n$ & \cellcolor{red!20}{$d_1$-$p_1$}    & \cellcolor{cyan!20}$p_1$-$d_1$  &\cellcolor{red!20} $p_2$-$d_2$   &\cellcolor{cyan!20} $d_2$-$p_2$ & Avg. &\cellcolor{red!20}Same & \cellcolor{cyan!20}Rev.  & Avg. \\
\midrule
For+Per & 1 & 98.67 & 25.76 & 75.15 & 16.00 & 53.90 & 91.0 & 7.0 & 49.0 \\
For+Per & 2 & \textbf{100.00} & 31.85 & 76.34 & 49.00 & 64.30 & \textbf{95.0} & 23.0 & 59.0 \\
For+Per & 3 & \underline {99.67} & 35.26 & \underline {80.51} & 74.33 & 72.44 & 90.0 & {30.0} & 60.0 \\
For+Per & 4 & 99.33 & \underline {40.38} & {78.71} & \textbf{93.00} & \underline {77.86} & \textbf{95.0} & 28.0 & {61.5} \\
For+Per & 5 & \underline {99.67} & 37.42 & \textbf{80.68} & \underline {90.33} & 77.03 & {92.0} & \underline {49.0} & \underline {70.5} \\
\rowcolor{Gray}
w/ sem &  & \underline {99.67} & \textbf{53.27} & 76.92 & 88.33 & \textbf{79.55} & \underline {93.0} & \textbf{78.0} & \textbf{85.5} \\
\midrule
Bi & 1 & 99.00 & 23.16 & 75.27 & 4.33 & 50.44 & 89.0 & 10.0 & 49.5 \\
Bi & 2 & \underline {99.67} & 27.62 & 74.07 & 55.67 & 64.26 & 92.0 & 15.0 & 53.5 \\
Bi & 3 & \underline {99.67} & 34.12 & 74.28 & 72.00 & 70.02 & \textbf{95.0} & 23.0 & 59.0 \\
Bi & 4 & 99.33 & 38.92 & \textbf{76.65} & 89.00 & 75.98 & 92.0 & 18.0 & 55.0 \\
Bi & 5 & 99.00 & \underline {39.70} & 74.56 & \textbf{92.67} & \underline {76.48} & \underline {93.0} & \underline {34.0} & \underline {63.5} \\
\rowcolor{Gray}
w/ sem &  & \underline {99.67} & \textbf{57.52} & \underline {76.25} & \underline {90.00} & \textbf{80.86} & 91.0 & \textbf{81.0} & \textbf{86.0} \\
\midrule
Tri & 1 & 98.67 & 22.99 & 68.98 & 18.33 & 52.24 & 81.0 & 10.0 & 45.5 \\
Tri & 2 & \textbf {99.67} & 29.49 & 73.00 & 58.67 & 65.21 & 84.0 & 19.0 & 51.5 \\
Tri & 3 & \underline {99.00} & 37.41 & 75.56 & 89.67 & 75.41 & \textbf{91.0} & 32.0 & 61.5 \\
Tri & 4 & \textbf {99.67} & \underline {49.60} & 73.97 & \underline {95.33} & \underline {79.64} & 88.0 & 20.0 & 54.0 \\
Tri & 5 & 97.00 & 44.04 & \underline {76.06} & \textbf{96.67} & 78.44 & \underline {90.0} & \underline {52.0} & \underline {71.0} \\
\rowcolor{Gray}
w/ sem &  & \textbf {99.67} & \textbf{72.12} & \textbf{80.24} & \underline {95.33} & \textbf{86.84} & \underline {90.0} & \textbf{87.0} & \textbf{88.5}\\
\bottomrule
\end{tabular}
}
\caption{Results of SPT and chunks of specified length under different permutation strategy on two datasets. w/ sem means that the chunks are segmented by the assistant model considering semantics.
}
\label{tab:exp-person-semantic}
\end{table}

\vspace{-0.5cm}
\paragraph{Why do we need semantics?}
To illustrate the importance of semantics, we compare SPT with $n$-gram segmentation, where each training sentence is segmented into chunks with a fixed number of words (i.e., $n$). 
We report the results of the Person Description and QA dataset, ranging from uni-gram to 5-gram, or segmented by semantics, under different setting in Table~\ref{tab:exp-person-semantic}.

We observe under different permutation strategies during the training stage, the introduction of semantic segmentation results in an improvement in reversal questions on both datasets. 
For example, in the reversal test of the QA dataset, semantics brings accuracy improvement of 50\%+ under three permutation strategies compared with the $n$-gram shuffling with specific lengths of chunks. 
In addition, the best-fit $n$ varies from dataset. Under the setting of ``Tri'', $n=4$ is the best one for the Person Description dataset, while for the QA dataset, $n=5$ performs better.
Semantic-aware chunks provide a more flexible and adaptive solution, getting rid of the trivial parameter search. 

\paragraph{Permutation probability}
The ratio of re-ordering selected from \{original, permuting, reverse\} can be adjusted as required. By default, we employ the probability of $\frac{1}{3}$ for each order.
We vary the probability ratio to investigate the effect of the ratio. 
The results are reported in Table~\ref{tab:exp-trade}.

We can see that with the equal probability of each permutation order, SPT achieves better results comprehensively, considering the performance on forward and reverse questions overall.

\begin{table}[!htb]
\resizebox{\columnwidth}{!}{
\begin{tabular}{lll|lllll|lll}
\toprule
\multicolumn{3}{c|}{\textbf{Probability}}& \multicolumn{5}{c|}{\textbf{Person Description}} & \multicolumn{3}{c}{\textbf{QA}} \\
For & Per & Rev &  \cellcolor{red!20}{$d_1$-$p_1$}    & \cellcolor{cyan!20}$p_1$-$d_1$  &\cellcolor{red!20} $p_2$-$d_2$   &\cellcolor{cyan!20} $d_2$-$p_2$  & Avg. & 
\cellcolor{red!20}Same & \cellcolor{cyan!20}Rev.  & Avg. \\
\midrule
1.00 & 0 & 0 & \textbf{100} & 20.28 & 79.02 & 1.67  & 50.24 & \textbf{100} & 3 & 51.5 \\
0.5 & 0.25 & 0.25 & 99.67 & 60.78 & 77.52 & 96.33 & 83.58 &  91 & 80 & 85.5 \\
\rowcolor{Gray}
$1/3$ & $1/3$ & $1/3$ & 99.67 & \textbf{72.08} & \textbf{80.24} & 95.67 & \textbf{86.92} &  90 & 87 & \textbf{88.5} \\
0.25 & 0.25 & 0.5 & 99.67 & 61.63 & 75.69 &\textbf{98.67} & 83.92 & 81 & \textbf{89} & 85 \\
\bottomrule
\end{tabular}
}
\caption{Results of SPT under different probability ratios of re-ordering (forward (i.e., original), permute, reverse) on Person Description and QA dataset. 
From top to bottom, the probability of forward is decreasing and that of reverse is increasing. The row in gray is our default setting.
}
\label{tab:exp-trade}
\end{table}

\vspace{-0.5cm}
\section{Conclusion}
In this work, we conduct in-depth evaluations to analyze the root cause of the reversal curse on causal LLMs.
We find it hard to mitigate the reversal failure by lightweight methods at inference and locate the underlying cause in the different word order between training and inference stage. 
Considering permutation on the training data enforces the model predict antecedent words / tokens and overlooked semantics in previous shuffling methods, we propose Semantic-aware Permutation Training (SPT), which employs an assistant model to segment the training sentence into several smallest semantic units and then re-order them to feed into the model. 
Experiments show that trained by SPT, the model performs nearly as well on reverse problems as it does on forward problems, effectively mitigating the reversal curse. Moreover, SPT significantly advances the existing works. 
We hope our research will shed light on further explorations of LLMs.

\section*{Limitations}
This work analyzes the root cause of the reversal curse in depth and proposes an effective method of SPT to mitigate the challenge. 
Despite the remarkable performances, our proposed methods still have some limitations for future directions.
\textbf{Firstly}, it is recognized that the ability to understand bi-directional MLMs is considered stronger than that of autoregressive ones.
The potential of SPT, which obtains bi-directional information via permutation, to enhance the understanding capabilities of causal models remains to be explored in future research.
\textbf{Secondly}, 
Our findings inspire future research in the in-depth analysis and exploration of LLMs, encouraging innovative applications.

\section*{Ethics Statement}
All the experiments are conducted on existing datasets used in previous public related papers. We keep fair and honest in our analysis of experimental results, and our work does not harm anyone. We will make our code open-sourced for further explorations.
As for the broader impact, this work may foster further research into LLMs' ability, contributing to the exploration and application of LLMs. 
Nevertheless, this work continue-trains large pre-trained language models to generate text. 
Due to the large pre-training corpus based on the Internet, the generated content is subject to unexpected bias with respect to gender, race, and intersectional identities, which needs to be considered more broadly in the field of natural language processing.
\label{sec:bibtex}

\bibliography{anthology,custom}

\begin{thebibliography}{28}
\expandafter\ifx\csname natexlab\endcsname\relax\def\natexlab#1{#1}\fi

\bibitem[{Abdou et~al.(2022)Abdou, Ravishankar, Kulmizev, and S{\o}gaard}]{abdou2022word}
Mostafa Abdou, Vinit Ravishankar, Artur Kulmizev, and Anders S{\o}gaard. 2022.
\newblock Word order does matter and shuffled language models know it.
\newblock In \emph{Proceedings of the 60th Annual Meeting of the Association for Computational Linguistics (Volume 1: Long Papers)}, pages 6907--6919.

\bibitem[{Allen{-}Zhu and Li(2023)}]{DBLP:journals/corr/abs-2309-14402}
Zeyuan Allen{-}Zhu and Yuanzhi Li. 2023.
\newblock \href {https://doi.org/10.48550/ARXIV.2309.14402} {Physics of language models: Part 3.2, knowledge manipulation}.
\newblock \emph{CoRR}, abs/2309.14402.

\bibitem[{Berglund et~al.(2023)Berglund, Tong, Kaufmann, Balesni, Stickland, Korbak, and Evans}]{DBLP:journals/corr/abs-2309-12288}
Lukas Berglund, Meg Tong, Max Kaufmann, Mikita Balesni, Asa~Cooper Stickland, Tomasz Korbak, and Owain Evans. 2023.
\newblock \href {https://doi.org/10.48550/ARXIV.2309.12288} {The reversal curse: Llms trained on "a is b" fail to learn "b is a"}.
\newblock \emph{CoRR}, abs/2309.12288.

\bibitem[{Cao et~al.(2023)Cao, Kojima, Matsuo, and Iwasawa}]{cao2023unnatural}
Qi~Cao, Takeshi Kojima, Yutaka Matsuo, and Yusuke Iwasawa. 2023.
\newblock Unnatural error correction: Gpt-4 can almost perfectly handle unnatural scrambled text.
\newblock In \emph{Proceedings of the 2023 Conference on Empirical Methods in Natural Language Processing}, pages 8898--8913.

\bibitem[{Chiang et~al.(2023)Chiang, Li, Lin, Sheng, Wu, Zhang, Zheng, Zhuang, Zhuang, Gonzalez, Stoica, and Xing}]{vicuna2023}
Wei-Lin Chiang, Zhuohan Li, Zi~Lin, Ying Sheng, Zhanghao Wu, Hao Zhang, Lianmin Zheng, Siyuan Zhuang, Yonghao Zhuang, Joseph~E. Gonzalez, Ion Stoica, and Eric~P. Xing. 2023.
\newblock \href {https://lmsys.org/blog/2023-03-30-vicuna/} {Vicuna: An open-source chatbot impressing gpt-4 with 90\%* chatgpt quality}.

\bibitem[{Devlin et~al.(2019)Devlin, Chang, Lee, and Toutanova}]{DBLP:conf/naacl/DevlinCLT19}
Jacob Devlin, Ming{-}Wei Chang, Kenton Lee, and Kristina Toutanova. 2019.
\newblock \href {https://doi.org/10.18653/V1/N19-1423} {{BERT:} pre-training of deep bidirectional transformers for language understanding}.
\newblock In \emph{Proceedings of the 2019 Conference of the North American Chapter of the Association for Computational Linguistics: Human Language Technologies, {NAACL-HLT} 2019, Minneapolis, MN, USA, June 2-7, 2019, Volume 1 (Long and Short Papers)}, pages 4171--4186. Association for Computational Linguistics.

\bibitem[{Du et~al.(2022)Du, Qian, Liu, Ding, Qiu, Yang, and Tang}]{du2022glm}
Zhengxiao Du, Yujie Qian, Xiao Liu, Ming Ding, Jiezhong Qiu, Zhilin Yang, and Jie Tang. 2022.
\newblock Glm: General language model pretraining with autoregressive blank infilling.
\newblock In \emph{Proceedings of the 60th Annual Meeting of the Association for Computational Linguistics (Volume 1: Long Papers)}, pages 320--335.

\bibitem[{Gao et~al.(2020)Gao, Biderman, Black, Golding, Hoppe, Foster, Phang, He, Thite, Nabeshima, Presser, and Leahy}]{pile}
Leo Gao, Stella Biderman, Sid Black, Laurence Golding, Travis Hoppe, Charles Foster, Jason Phang, Horace He, Anish Thite, Noa Nabeshima, Shawn Presser, and Connor Leahy. 2020.
\newblock The {P}ile: An 800gb dataset of diverse text for language modeling.
\newblock \emph{arXiv preprint arXiv:2101.00027}.

\bibitem[{Gao et~al.(2023)Gao, Madaan, Zhou, Alon, Liu, Yang, Callan, and Neubig}]{gao2023pal}
Luyu Gao, Aman Madaan, Shuyan Zhou, Uri Alon, Pengfei Liu, Yiming Yang, Jamie Callan, and Graham Neubig. 2023.
\newblock Pal: Program-aided language models.
\newblock In \emph{International Conference on Machine Learning}, pages 10764--10799. PMLR.

\bibitem[{Grosse et~al.(2023)Grosse, Bae, Anil, Elhage, Tamkin, Tajdini, Steiner, Li, Durmus, Perez, Hubinger, Lukosiute, Nguyen, Joseph, McCandlish, Kaplan, and Bowman}]{DBLP:journals/corr/abs-2308-03296}
Roger~B. Grosse, Juhan Bae, Cem Anil, Nelson Elhage, Alex Tamkin, Amirhossein Tajdini, Benoit Steiner, Dustin Li, Esin Durmus, Ethan Perez, Evan Hubinger, Kamile Lukosiute, Karina Nguyen, Nicholas Joseph, Sam McCandlish, Jared Kaplan, and Samuel~R. Bowman. 2023.
\newblock \href {https://doi.org/10.48550/ARXIV.2308.03296} {Studying large language model generalization with influence functions}.
\newblock \emph{CoRR}, abs/2308.03296.

\bibitem[{Gupta et~al.(2021)Gupta, Kvernadze, and Srikumar}]{gupta2021bert}
Ashim Gupta, Giorgi Kvernadze, and Vivek Srikumar. 2021.
\newblock Bert \& family eat word salad: Experiments with text understanding.
\newblock In \emph{Proceedings of the AAAI conference on artificial intelligence}, volume~35, pages 12946--12954.

\bibitem[{Joshi et~al.(2020)Joshi, Chen, Liu, Weld, Zettlemoyer, and Levy}]{joshi2020spanbert}
Mandar Joshi, Danqi Chen, Yinhan Liu, Daniel~S Weld, Luke Zettlemoyer, and Omer Levy. 2020.
\newblock Spanbert: Improving pre-training by representing and predicting spans.
\newblock \emph{Transactions of the Association for Computational Linguistics}, 8:64--77.

\bibitem[{Kojima et~al.(2022)Kojima, Gu, Reid, Matsuo, and Iwasawa}]{kojima2022large}
Takeshi Kojima, Shixiang~Shane Gu, Machel Reid, Yutaka Matsuo, and Yusuke Iwasawa. 2022.
\newblock Large language models are zero-shot reasoners.
\newblock \emph{Advances in neural information processing systems}, 35:22199--22213.

\bibitem[{Lewis et~al.(2020)Lewis, Liu, Goyal, Ghazvininejad, Mohamed, Levy, Stoyanov, and Zettlemoyer}]{lewis2020bart}
Mike Lewis, Yinhan Liu, Naman Goyal, Marjan Ghazvininejad, Abdelrahman Mohamed, Omer Levy, Veselin Stoyanov, and Luke Zettlemoyer. 2020.
\newblock Bart: Denoising sequence-to-sequence pre-training for natural language generation, translation, and comprehension.
\newblock In \emph{Proceedings of the 58th Annual Meeting of the Association for Computational Linguistics}, pages 7871--7880.

\bibitem[{Li et~al.(2023)Li, Zhang, Zhao, Yang, and Yang}]{DBLP:journals/corr/abs-2307-00360}
Zuchao Li, Shitou Zhang, Hai Zhao, Yifei Yang, and Dongjie Yang. 2023.
\newblock \href {https://doi.org/10.48550/ARXIV.2307.00360} {Batgpt: {A} bidirectional autoregessive talker from generative pre-trained transformer}.
\newblock \emph{CoRR}, abs/2307.00360.

\bibitem[{Lv et~al.(2023)Lv, Zhang, Xie, Tu, Chen, Wen, and Yan}]{DBLP:journals/corr/abs-2311-07468}
Ang Lv, Kaiyi Zhang, Shufang Xie, Quan Tu, Yuhan Chen, Ji{-}Rong Wen, and Rui Yan. 2023.
\newblock \href {https://doi.org/10.48550/ARXIV.2311.07468} {Are we falling in a middle-intelligence trap? an analysis and mitigation of the reversal curse}.
\newblock \emph{CoRR}, abs/2311.07468.

\bibitem[{Ma et~al.(2023)Ma, Gu, Ling, Liu, and Liu}]{DBLP:journals/corr/abs-2310-10322}
Jun{-}Yu Ma, Jia{-}Chen Gu, Zhen{-}Hua Ling, Quan Liu, and Cong Liu. 2023.
\newblock \href {https://doi.org/10.48550/ARXIV.2310.10322} {Untying the reversal curse via bidirectional language model editing}.
\newblock \emph{CoRR}, abs/2310.10322.

\bibitem[{OpenAI(2023)}]{DBLP:journals/corr/abs-2303-08774}
OpenAI. 2023.
\newblock \href {https://doi.org/10.48550/ARXIV.2303.08774} {{GPT-4} technical report}.
\newblock \emph{CoRR}, abs/2303.08774.

\bibitem[{Ouyang et~al.(2022)Ouyang, Wu, Jiang, Almeida, Wainwright, Mishkin, Zhang, Agarwal, Slama, Ray et~al.}]{ouyang2022training}
Long Ouyang, Jeffrey Wu, Xu~Jiang, Diogo Almeida, Carroll Wainwright, Pamela Mishkin, Chong Zhang, Sandhini Agarwal, Katarina Slama, Alex Ray, et~al. 2022.
\newblock Training language models to follow instructions with human feedback.
\newblock \emph{Advances in Neural Information Processing Systems}, 35:27730--27744.

\bibitem[{Papineni et~al.(2002)Papineni, Roukos, Ward, and Zhu}]{papineni-etal-2002-bleu}
Kishore Papineni, Salim Roukos, Todd Ward, and Wei-Jing Zhu. 2002.
\newblock \href {https://doi.org/10.3115/1073083.1073135} {{B}leu: a method for automatic evaluation of machine translation}.
\newblock In \emph{Proceedings of the 40th Annual Meeting of the Association for Computational Linguistics}, pages 311--318, Philadelphia, Pennsylvania, USA. Association for Computational Linguistics.

\bibitem[{Pham et~al.(2021)Pham, Bui, Mai, and Nguyen}]{DBLP:conf/acl/PhamBMN21}
Thang~M. Pham, Trung Bui, Long Mai, and Anh Nguyen. 2021.
\newblock \href {https://doi.org/10.18653/V1/2021.FINDINGS-ACL.98} {Out of order: How important is the sequential order of words in a sentence in natural language understanding tasks?}
\newblock In \emph{Findings of the Association for Computational Linguistics: {ACL/IJCNLP} 2021, Online Event, August 1-6, 2021}, volume {ACL/IJCNLP} 2021 of \emph{Findings of {ACL}}, pages 1145--1160. Association for Computational Linguistics.

\bibitem[{Rozi{\`{e}}re et~al.(2023)Rozi{\`{e}}re, Gehring, Gloeckle, Sootla, Gat, Tan, Adi, Liu, Remez, Rapin, Kozhevnikov, Evtimov, Bitton, Bhatt, Canton{-}Ferrer, Grattafiori, Xiong, D{\'{e}}fossez, Copet, Azhar, Touvron, Martin, Usunier, Scialom, and Synnaeve}]{DBLP:journals/corr/abs-2308-12950}
Baptiste Rozi{\`{e}}re, Jonas Gehring, Fabian Gloeckle, Sten Sootla, Itai Gat, Xiaoqing~Ellen Tan, Yossi Adi, Jingyu Liu, Tal Remez, J{\'{e}}r{\'{e}}my Rapin, Artyom Kozhevnikov, Ivan Evtimov, Joanna Bitton, Manish Bhatt, Cristian Canton{-}Ferrer, Aaron Grattafiori, Wenhan Xiong, Alexandre D{\'{e}}fossez, Jade Copet, Faisal Azhar, Hugo Touvron, Louis Martin, Nicolas Usunier, Thomas Scialom, and Gabriel Synnaeve. 2023.
\newblock \href {https://doi.org/10.48550/ARXIV.2308.12950} {Code llama: Open foundation models for code}.
\newblock \emph{CoRR}, abs/2308.12950.

\bibitem[{Sinha et~al.(2021{\natexlab{a}})Sinha, Jia, Hupkes, Pineau, Williams, and Kiela}]{sinha2021masked}
Koustuv Sinha, Robin Jia, Dieuwke Hupkes, Joelle Pineau, Adina Williams, and Douwe Kiela. 2021{\natexlab{a}}.
\newblock Masked language modeling and the distributional hypothesis: Order word matters pre-training for little.
\newblock In \emph{Proceedings of the 2021 Conference on Empirical Methods in Natural Language Processing}, pages 2888--2913.

\bibitem[{Sinha et~al.(2021{\natexlab{b}})Sinha, Parthasarathi, Pineau, and Williams}]{sinha2021unnatural}
Koustuv Sinha, Prasanna Parthasarathi, Joelle Pineau, and Adina Williams. 2021{\natexlab{b}}.
\newblock Unnatural language inference.
\newblock In \emph{Proceedings of the 59th Annual Meeting of the Association for Computational Linguistics and the 11th International Joint Conference on Natural Language Processing (Volume 1: Long Papers)}, pages 7329--7346.

\bibitem[{Song et~al.(2019)Song, Tan, Qin, Lu, and Liu}]{song2019mass}
Kaitao Song, Xu~Tan, Tao Qin, Jianfeng Lu, and Tie-Yan Liu. 2019.
\newblock Mass: Masked sequence to sequence pre-training for language generation.
\newblock In \emph{International Conference on Machine Learning}, pages 5926--5936. PMLR.

\bibitem[{Song et~al.(2020)Song, Tan, Qin, Lu, and Liu}]{song2020mpnet}
Kaitao Song, Xu~Tan, Tao Qin, Jianfeng Lu, and Tie-Yan Liu. 2020.
\newblock Mpnet: Masked and permuted pre-training for language understanding.
\newblock \emph{Advances in Neural Information Processing Systems}, 33:16857--16867.

\bibitem[{Touvron et~al.(2023)Touvron, Lavril, Izacard, Martinet, Lachaux, Lacroix, Rozi{\`{e}}re, Goyal, Hambro, Azhar, Rodriguez, Joulin, Grave, and Lample}]{DBLP:journals/corr/abs-2302-13971}
Hugo Touvron, Thibaut Lavril, Gautier Izacard, Xavier Martinet, Marie{-}Anne Lachaux, Timoth{\'{e}}e Lacroix, Baptiste Rozi{\`{e}}re, Naman Goyal, Eric Hambro, Faisal Azhar, Aur{\'{e}}lien Rodriguez, Armand Joulin, Edouard Grave, and Guillaume Lample. 2023.
\newblock \href {https://doi.org/10.48550/ARXIV.2302.13971} {Llama: Open and efficient foundation language models}.
\newblock \emph{CoRR}, abs/2302.13971.

\bibitem[{Yang et~al.(2019)Yang, Dai, Yang, Carbonell, Salakhutdinov, and Le}]{yang2019xlnet}
Zhilin Yang, Zihang Dai, Yiming Yang, Jaime Carbonell, Russ~R Salakhutdinov, and Quoc~V Le. 2019.
\newblock Xlnet: Generalized autoregressive pretraining for language understanding.
\newblock \emph{Advances in neural information processing systems}, 32.

\end{thebibliography}
\bibliographystyle{acl_natbib}

\appendix
\section{Experimental Settings}
\subsection{Prompts}
\label{sec:prompt}
When querying the assistant model to segment the sentence into semantic-aware chunks, we use the few-shot demonstration shown in Figure~\ref{fig:demon-sep}. 

For experiments on the Celebrity Relation dataset, we prepend few-shot demonstrations at inference, either w/ (Figure~\ref{fig:demon-cot}) or w/o CoT (Figure~\ref{fig:demon-def}).

\begin{center}
 \begin{tcbitemize}[raster columns=1,raster equal height,
colframe=black!75!black,colback=black!-15!white,fonttitle=\bfseries]
\tcbitem[squeezed title={Demonstration for segmentation.}]
\fontsize{10}{13}\selectfont{
\small
Below is a converation with a helpful and terse assistant. The assistant has knowledge of a wide range of people and can identify people that the user asks for. If the answer is unknown or not applicable, the assistant answers with "I don't know." \\

Q: Who is Elon Musk's mother?\\
A: Maye Musk. \\

Q: Who is Malia Obama's father? \\
A: Barack Obama. \\

Q: Who is Jennifer Lawrence's mother?\\
A: Karen Lawrence. \\

Q: Who is Aaron Taylor-Johnson's mother? \\
A: Sarah Johnson. \\

Q: Who is Chris Hemsworth's father? \\
A: Craig Hemsworth. \\

Q: Who is Sasha Calle's mother? \\
A:
}
\end{tcbitemize}
\captionof{figure}{Demonstration used for celebrity relation dataset at inference (w/o CoT).}
\label{fig:demon-def}

\end{center}

\begin{center}
 \begin{tcbitemize}[raster columns=1,raster equal height,
colframe=black!75!black,colback=black!-15!white,fonttitle=\bfseries]
\tcbitem[squeezed title={Demonstration for segmentation.}]
\fontsize{10}{13}\selectfont{
\small
Below is a converation with a helpful and terse assistant. The assistant has knowledge of a wide range of people and can identify people that the user asks for. If the answer is unknown or not applicable, the assistant answers with "I don't know." \\

Q: Who is Elon Musk's mother?\\
A: Elon Musk's mother is Maye Musk. Maye Musk is Elon Musk's mother.\\

Q: Who is Malia Obama's father?\\
A: Malia Obama's father is Barack Obama. Barack Obama is Malia Obama's father.\\

Q: Who is Jennifer Lawrence's mother?\\
A: Jennifer Lawrence's mother is Karen Lawrence. Karen Lawrence is Jennifer Lawrence's mother.\\

Q: Who is Aaron Taylor-Johnson's mother?\\
A: Aaron Taylor-Johnson's mother is Sarah Johnson. Sarah Johnson is Aaron Taylor-Johnson's mother.\\

Q: Who is Chris Hemsworth's father?\\
A: Chris Hemsworth's father is Craig Hemsworth. Craig Hemsworth is Chris Hemsworth's father.\\

Q: Who is Sasha Calle's mother? \\
A:}
\end{tcbitemize}
\captionof{figure}{An example CoT demonstration used for Celebrity Relation dataset at inference for model M1 when tested on question Q1 (w/ CoT, corresponding to Table~\ref{tab:app-cot-demon}).}
\label{fig:demon-cot}
\end{center}

\begin{center}
 \begin{tcbitemize}[raster columns=1,raster equal height,
colframe=black!75!black,colback=black!-15!white,fonttitle=\bfseries]
\tcbitem[squeezed title={Demonstration for segmentation.}]
\fontsize{10}{13}\selectfont{
A chat between a curious user and an artificial intelligence assistant. The assistant gives helpful, detailed, and polite answers to the user's questions.  \\ 

\textbf{USER}: \\
Segment the input sentence into the smallest semantic units using [SEP] token, and make sure that each unit contains actual meaning. Note that there should be at least one [SEP] token. Do not delete or add any other words and not put the token at the end of the sentence.\\

\textbf{Input}: You can play “Survival of the Tastiest” on Android, and on the web. Playing on the web works, but you have to simulate multi-touch for table moving and that can be a bit confusing.\\
\textbf{Output}: You can play [SEP] "Survival of the Tastiest" [SEP] on Android, [SEP] and on the web. [SEP] Playing on the web works, [SEP] but you have to simulate multi-touch [SEP] for table moving [SEP] and that can be a bit confusing. \\

\textbf{Input}: Pastas used in the game. Unfortunately, the macs where never used \\
\textbf{Output}: Pastas [SEP] used in the game. [SEP] Unfortunately, the macs where never used \\

\textbf{Input}: At the same time, I do know it was the right thing to do given the timeframe. \\
\textbf{Output}: At the same time, [SEP] I do know [SEP] it was the right thing [SEP] to do given the timeframe. \\

\textbf{Input}: Never shy about being the best-selling author of the self-help book, "Unleashing Your Inner Superhero.", Lacey Donnelly lives life on their own terms. \\
\textbf{Output}: Never shy [SEP] about being the best-selling author [SEP] of the self-help book, [SEP] "Unleashing Your Inner Superhero.", [SEP] Lacey Donnelly lives life [SEP] on their own terms.  \\

\textbf{Input}: <prompt>  \\
\textbf{Output}: }
\end{tcbitemize}
\captionof{figure}{Demonstration used for segmenting the sentence into smallest semantic units. The input examples are randomly sampled from Pile~\citep{pile}.}
\label{fig:demon-sep}
\end{center}

\begin{table}[t]
\resizebox{\columnwidth}{!}{
    \begin{tabular}{ll|l}
    \toprule
    \textbf{Model} & \textbf{Question} & \textbf{Template} \\
    \midrule
M1 & Q1 & A's father is B.  B is A's father. \\
M1 & Q2 & A's father is B.  A's father is B. \\
M1 & Q3 & A's father is B. B's child is A. \\
M1 & Q4 & A's father is B.  A's child is B. \\
M1 & Q5 & A's father is B.  B is A's father. \\
M1 & Q6 & A's father is B. A's father is B. \\
M1 & Q7 & A's father is B.  B's child is A. \\
M1 & Q8 & A's father is B.  A is B's child. \\
\midrule
M2 & Q1 & A is B's child. B is A's father. \\
M2 & Q2 & A is B's child. A's father is B. \\
M2 & Q3 & A is B's child. B's child is A. \\
M2 & Q4 & A is B's child. A's child is B. \\
M2 & Q5 & A is B's child. B is A's father. \\
M2 & Q6 & A is B's child. A's father is B. \\
M2 & Q7 & A is B's child. B's child is A. \\
M2 & Q8 & A is B's child. A is B's child. \\
\midrule
M3 & Q1 & B is A's father. B is A's father. \\
M3 & Q2 & B is A's father. A's father is B. \\
M3 & Q3 & B is A's father. B's child is A. \\
M3 & Q4 & B is A's father. A's child is B. \\
M3 & Q5 & B is A's father. B is A's father. \\
M3 & Q6 & B is A's father. A's father is B. \\
M3 & Q7 & B is A's father. B's child is A. \\
M3 & Q8 & B is A's father. A is B's child. \\
\midrule
M4 & Q1 & B's child is A. B is A's father. \\
M4 & Q2 & B's child is A. A's father is B. \\
M4 & Q3 & B's child is A. B's child is A. \\
M4 & Q4 & B's child is A. A's child is B. \\
M4 & Q5 & B's child is A. B is A's father. \\
M4 & Q6 & B's child is A. A's father is B. \\
M4 & Q7 & B's child is A. B's child is A. \\
M4 & Q8 & B's child is A. A is B's child.  \\
    \bottomrule
    \end{tabular}
    }
\caption{Chain-of-Thought reasoning path of eight testing questions for four models in the Celebrity Relation dataset.}
\label{tab:app-cot-demon}
\end{table}
\begin{table}[htb!]
\resizebox{\columnwidth}{!}{
\begin{tabular}{l|lll}
\toprule
Hyper-parameters & Celebrity Relation & Person Description & QA \\
\midrule
Warmup Ratio & 0.03 & 0.03 & 0.03 \\
Weight Decay & 0 & 0 & 0 \\
Learning Rate & 2e-5 & 2e-5 & 2e-5 \\
Batch Size & 128 & 128 & 128 \\
Epoch & 30 & 10 & 20 \\
Epoch* & - & 3 & 20 \\
\bottomrule
\end{tabular}
}
\caption{Hyper-parameters for SPT of different datasets. * refers to the setting used in Section \S \ref{sec:discussion}. }
\label{tab:app-param}
\end{table}
\subsection{Hyper Parameters}
\label{sec:param}
Hyper-parameters for all experiments can be found in Table~\ref{tab:app-param}. We conduct our experiments on open-sourced LLMs with the code base of Stanford Alpaca\footnote{\url{https://github.com/tatsu-lab/stanford\_alpaca}}. 
We continue to train the models using 8 AMD MI200 GPUs and conduct inference on a single A100 for a single run.

\end{document}